\pdfoutput=1

\documentclass[11pt]{article}

\usepackage[preprint]{acl}
\usepackage{multirow}
\usepackage{booktabs}
\usepackage{times}
\usepackage{latexsym}
\usepackage{array}
\usepackage{graphicx} 
\usepackage{afterpage}
\usepackage{lipsum} 
\usepackage{hyperref}

\usepackage[T1]{fontenc}

\usepackage[utf8]{inputenc}

\usepackage{microtype}
\usepackage{algorithm}
\usepackage{algpseudocode}

\usepackage{inconsolata}
\usepackage{amsmath}
\usepackage{graphicx}
\newcommand{\model}[1]{AnomalyGen}
\usepackage{xcolor} 

\title{Hazards in Daily Life? \\Enabling Robots to Proactively Detect and Resolve Anomalies}

\def\SP{~~}

\author{
\textmd{Zirui Song}$^{1,2 \ast}$,
\SP
\textmd{Guangxian Ouyang}$^3$\thanks{Equal contributions.}, 
\SP 
\textmd{Meng Fang}$^{4 \ast \dag}$,
\SP
\textmd{Hongbin Na}$^2$,
\SP 
\textmd{Zijing Shi}$^2$, \\
\SP
\textmd{Zhenhao Chen}$^{1}$,
\SP
\textmd{Yujie Fu}$^{3}$,
\SP
\textmd{Zeyu Zhang}$^{6}$
\SP
\textmd{Shiyu Jiang}$^{5}$
\SP
\textmd{Miao Fang}$^{3}$,
\SP
\textmd{Ling Chen}$^{2}$ \\
\SP 
\textmd{Xiuying Chen}$^{1 \thanks{Corresponding author}}$
\\[0.1125cm]
\normalsize 
$ ^1$ Mohamed bin Zayed University of Artificial Intelligence 
\normalsize \\
\normalsize
\SP ~~~
$ ^2 $ University of Technology Sydney
\SP ~~~
$ ^3 $ Northeastern University
\SP ~~~
$ ^4 $ University of Liverpool 
\normalsize
\\
\normalsize
\SP ~~~
$ ^5$  Johns Hopkins University
\normalsize 
\SP ~~~
$ ^6$ The Australian National University
\normalsize
\\
\\
}

\begin{document}
\maketitle
\begin{abstract}
Existing household robots have made significant progress in performing routine tasks, such as cleaning floors or delivering objects. 
However, a key limitation of these robots is their inability to recognize potential problems or dangers in home environments.
For example, a child may pick up and ingest medication that has fallen on the floor, posing a serious risk. 
We argue that household robots should proactively detect such hazards or anomalies within the home, and propose the task of \textit{anomaly scenario generation}.
We leverage foundational models instead of relying on manually labeled data to build simulated environments.
Specifically, we introduce a multi-agent brainstorming approach, where agents collaborate and generate diverse scenarios covering household hazards, hygiene management, and child safety. 
These textual task descriptions are then integrated with designed 3D assets to simulate realistic environments.
Within these constructed environments, the robotic agent learns the necessary skills to proactively discover and handle the proposed anomalies through task decomposition, and optimal learning approach selection. 
We demonstrate that our generated environment outperforms others in terms of task description and scene diversity, ultimately enabling robotic agents to better address potential household hazards. 

\end{abstract}

\section{Introduction}
The development of Vision-Language Models (VLMs) has significantly improved household robots' ability to interact with the physical world in a more human-like manner \cite{liu2024visual,liu2024improved,cai2023benchlmm,majumdar2024openeqa}. Among these models, the most popular paradigm for such robots is receiving instructions and performing corresponding operational tasks \cite{yang2024embodiedmultimodalagenttrained,driess2023palm,ahn2022can}.

However, a critical yet often overlooked scenario arises when no instructions are provided.
According to survey data, 31\% of cooking fires are caused by unattended equipment \cite{fire}. 
Meanwhile, unintentional injuries are the predominant cause of death among children, particularly those aged 1-14 years, encompassing incidents such as drowning, falls, and accidental poisonings \cite{safekids2022}. 
A lack of adequate supervision is often identified as a significant contributor to many of these fatalities, especially in cases involving younger children \cite{aap2006,childstats2023}.
It would greatly benefit humans if household robots could monitor whether stoves and other fire sources are properly turned off and detect potential hazards in the home that could lead to falls or accidental poisonings. 
Many of these fires and unintentional injuries could be prevented.
However, to the best of our knowledge, such robotics have yet to be implemented.

\begin{figure}
    \centering
    \includegraphics[width=1\linewidth]{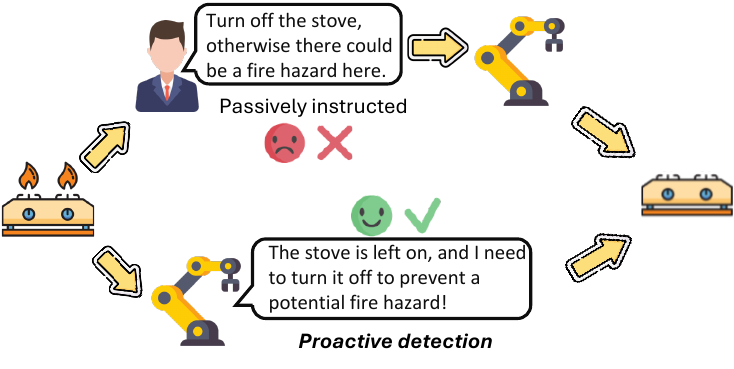}
    \caption{Comparison of passively instructed robots and our proactive detection robot. 
    Our paradigm creates benefits and convenience for safety, even in the absence of human presence.}
    \label{fig:intro}
\end{figure}

Hence, in this work, we propose \model{}, which can generate diverse \textit{anomaly settings covering household hazards, hygiene management, and child safety} in 3D simulation environments, enabling robots to develop proactive detection and problem-solving abilities, as shown in Figure~\ref{fig:intro}.
Specifically, we first devised a group brainstorming setting, where LLM-based agents collaborate to generate diverse and comprehensive anomaly scenarios.
The motivation comes from the observation that simply prompting an LLM to generate hazard scenarios results in repetitive and similar settings.
In contrast, group brainstorming in real-life meetings often leads to novel and creative ideas.
Based on these task settings, \model{} automatically constructs simulated anomalous scenes through carefully designed 3D asset retrieval, configuration, and scene setup steps.
Finally, \model{} guides household robots in developing detection and resolution abilities for handling anomalies. It reads textual descriptions of the simulated environment, including the 3D coordinates of assets, and automatically identifies potential anomalous tasks that require attention.
\model{} then decomposes the task into fine-grained sub-tasks and selects the most appropriate learning method for the household robot.
 In general, our \model{} leverages language-based approaches to bridge the domain gap between foundational models and robot interaction, enabling operations such as control inputs, operational trajectories, and physical interaction.

For the experiments, \model{} constructs 111 diverse and comprehensive anomaly scenes, with human evaluation showing high quality and automatic metrics demonstrating greater diversity compared to previous human-crafted robotic datasets.
Based on this simulation data, household robots are guided by \model{} to learn and demonstrate a variety of skills across tasks such as rigid and articulated object manipulation and legged locomotion, achieving a task completion rate of 83\%. 
Additionally, we conduct an error analysis highlighting the limitations of the current learning algorithm and VLM, identifying areas for future improvement and direction.

Our contributions can be summarized as follows:  
Firstly, we introduce \model{}, an unsupervised generative framework that enables household robots to autonomously detect and address anomalies without explicit instructions.  
Secondly, \model{} creates a 3D simulation environment with 111 diverse hazard scenarios, generated through a collaborative brainstorming mechanism, significantly enhancing task diversity compared to previous datasets.  
Thirdly, \model{} enables robots to autonomously identify anomalies, decompose tasks, and learn appropriate skills using an effective task decomposition and learning method with minimal human input.

\section{Related Work}
\subsection{Household Anomaly Detection}

Household robotics have seen significant advancements, with researchers developing various benchmarks for embodied AI agents to tackle household tasks in simulation. 
Behavior1K \cite{li2021igibson,shen2021igibson} introduced a benchmark for AI agents to complete 100 household activities in a simulated environment. Housekeep \cite{kant2022housekeeptidyingvirtualhouseholds} focuses on organizing homes by rearranging cluttered items, while TidyBot \cite{wu2023tidybot} emphasizes personalized household cleanup, aiming to understand and place items in their correct locations. However, limited work has been done on anomaly detection in household environments.
The only notable dataset addressing safety-related topics in this domain is SafetyDetect \cite{Mullen_2024}, which manually configured just seven distinct scenes and required substantial human effort for scene construction and data collection. 
Additionally, it is an image-based dataset, not a simulation.
In contrast, \model{} autonomously generates simulation environment and tasks without human intervention.

\subsection{Foundation Models in Robotics}

With the rapid development of foundation and generative models in multi-modal settings~\citep{poole2022dreamfusion, melas2023realfusion, touvron2023llama, driess2023palm, gpt4, liu2023audioldm, girdhar2023imagebind}, a growing body of research has begun to harness the capabilities of foundational models across various domains, such as visual imagination for skill execution~\citep{du2023learning}, and sub-task planning~\citep{ahn2022can, huang2022inner, lin2023text2motion}, among others.
Some recent works have also attempted to fully harness the potential of LLMs for robotic manipulation, such as using LLMs for reward function generation \citep{yu2023language, ma2023eureka}, and sub-task and trajectory generation \citep{ha2023scaling}. 
Additionally, GenSim explored LLM-based robotic instruction tasks, but it primarily focused on object manipulation on desktops with a limited set of 3D assets.
Gen2Sim \cite{katara2023gen2sim} extends the range of task types by generating instead of only retrieving new 3D assets.

\subsection{Simulation Environment}
VirtualHome~\cite{puig2018virtualhome} and Alfred~\cite{alfred} abstract physical interactions to concentrate on symbolic reasoning, yet they lack physical realism and a comprehensive scope of actions. Habitat~\cite{savva2019habitat}, employing 3D scans of real homes, focuses on navigation tasks~\cite{batra2020objectnav} but omits physics-based interactions. To augment physical realism, Habitat 2.0~\cite{homerobot} and iGibson~\cite{shen2021igibson, li2021igibson2} introduce realistic actions, interactions with environments, and object state simulations. Additionally, emerging simulation platforms such as ManiSkills~\cite{gu2023maniskill2}, TDW~\cite{tdwworld}, SoftGym~\cite{softgym}, and RFUniverse~\cite{fu2022rfuniverse} emphasize physical realism but still fall short on task diversity. To enrich task variety, several works have explored language-conditioned tasks~\cite{raven, james2019rlbench, mees2022calvin}. 
\model{} focuses on constructing household anomaly scenes, which is an area that has not been covered in previous work.

\section{Method}
In this section, we demonstrate how our framework utilizes advanced generative models to automatically create anomalous scenarios and task-related data, as shown in Figure \ref{fig:Framework}. 

\subsection{Brainstormed Anamoly Task Proposal}
\label{brain}
The most intuitive way to obtain anomaly scenarios would be to prompt an LLM to generate a list.
However, in our preliminary experiments, the LLM tends to generate repetitive and lackluster scenarios, such as 
"move the scissors to a drawer", "put the scissors into the box" and "store scissors safely."
This lack of diversity limits the range of potential hazards, resulting in scenarios that are too similar to each other.

To address this, we propose group brainstorming, a round-based divergent thinking framework that allows multiple agents to build upon each other’s ideas. 
We also incorporate role-playing, where each agent adopts a unique perspective, encouraging a broader range of creative thoughts.
This collaborative process not only enhances the variety of scenarios but also improves the realism and complexity of the generated anomalies.

\begin{figure*}[t] 
    \centering
    \includegraphics[width=\textwidth,height=0.55\textwidth]{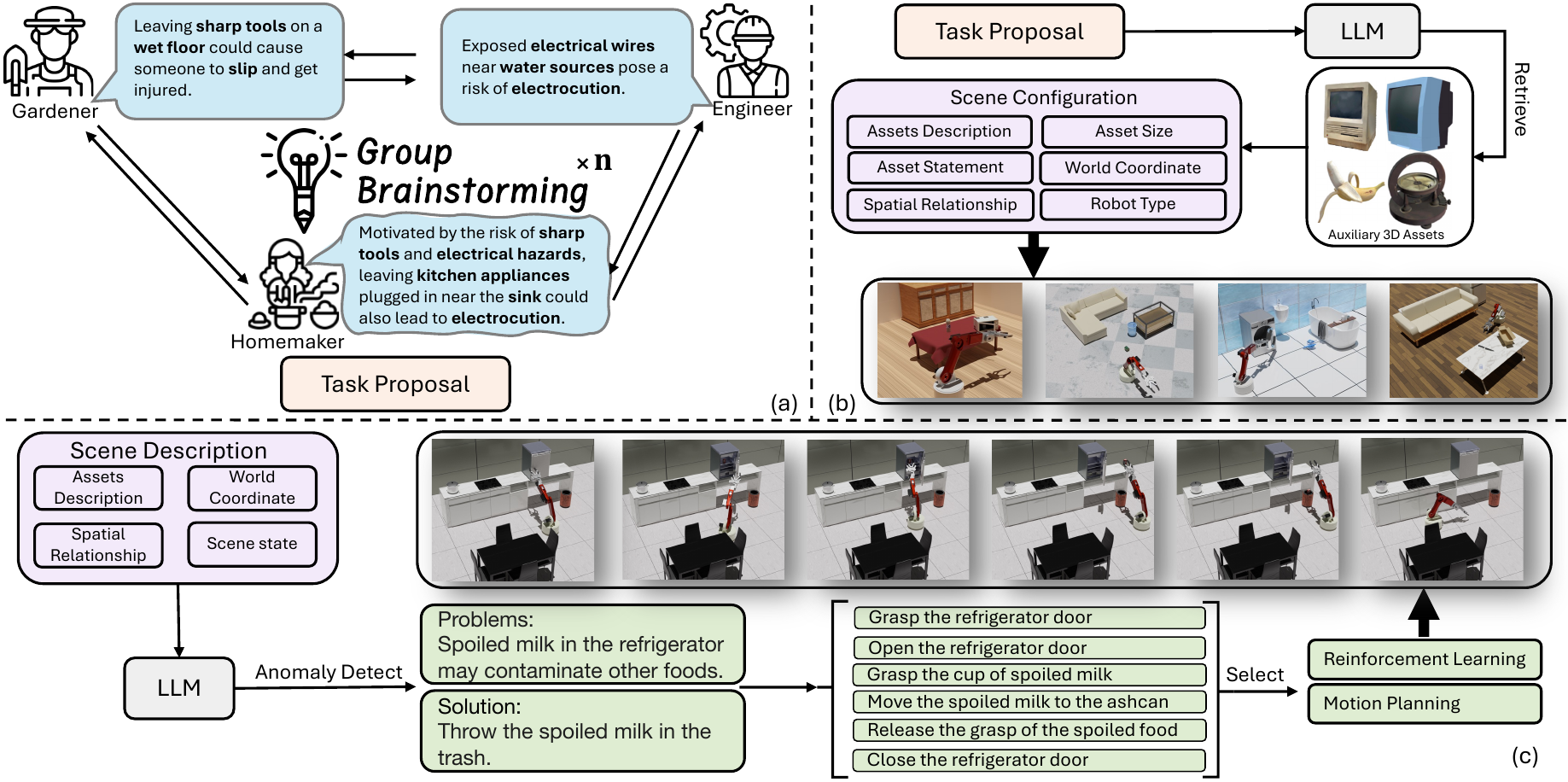} %
    \caption{\model{} includes 3 modules: a) Group Brainstorming, b) Anomalous Scenarios Generation, c) Proactive Anomaly Detection and Anomaly Task Learning.}
    \label{fig:Framework}
\end{figure*}

\textbf{Role-play Initialization Stage.}
To ensure that each agent considers different perspectives and approaches the task from various angles, we assign distinct roles to each agent. 
This is especially useful in households, where different professions face diverse hazard scenarios daily.
For example, a parent may focus on child safety, noticing hazards like sharp objects or unlocked cabinets, while a household maintenance worker may be more attuned to issues like electrical faults or fire hazards.
Our role-play list includes roles such as homemaker, household safety advisor, and educational consultant, with the full list in the Appendix \ref{appenix::Role list}.
Based on their assigned role, each agent randomly selects a \textit{target object} from an anomalous household asset list and proposes an initial anomaly scenario based on that object.
This anomalous object list was curated from a subset of the PartNet Mobility dataset~\cite{Xiang_2020_SAPIEN,wang2023robogen}. 
The detailed categories and directory of these anomalous object assets can be found in the Appendix \ref{appendix: assets stats}.


\textbf{Brainstorming Stage.} 
After each agent proposes its initial anomaly scenario, we facilitate multiple rounds of discussion to foster meaningful exchanges of ideas among the LLM agents.
Each agent takes the outputs from other agents in the previous rounds, combines them with its own character and thoughts, and proposes new tasks. 
Each agent is informed that they are part of a collaborative brainstorming session, where teamwork and diverse perspectives are key to generating creative and comprehensive hazard scenarios.
 The detailed prompt can be found in Appendix \ref{appendix: detail prompt}

For instance, as shown in Figure~\ref{fig:Framework}, the "gardener" and "engineer" agents propose a scenario where sharp tools left on a wet floor and exposed electrical wires near water pose significant safety risks. 
Motivated by the sharp tools mentioned by the gardener and the electrical hazards raised by the engineer, the "homemaker" suggests that leaving kitchen appliances plugged in near the sink could also lead to electrocution, expanding on the dangers of water-related hazards in the home.
This iterative dialogue mimics human brainstorming, where participants build on each other's ideas for more creative and comprehensive outcomes. 
By encouraging LLM agents to collaborate, we achieve greater variety and depth in the generated anomaly scenarios.

\subsection{3D Anomalous Scenarios Generation}
\label{verification}
The brainstormed ideas above are textual descriptions of the scenario, and our next step is to turn the text into vivid 3D environments.

\textbf{Auxiliary 3D Assets Retrieval.}
In the previous section, we compiled a list of 3D anomaly assets that serve as \textit{target objects} for the robots to manipulate. 
However, to realistically simulate real-world scenarios, focusing only on these target objects is not enough. 
We also need \textit{auxiliary surrounding objects} to construct realistic environments that mimic real-world object distributions. 
A straightforward approach to select these surrounding objects is to source them from Objaverse \cite{deitke2023objaverse}. However, this dataset is extensive, containing up to 800K items, and the item names are often too short or duplicated, making selection challenging.

To address this, we first query LLM to generate names and descriptions of objects relevant to the anomalous task, from a \textit{textual perspective}. 
We then employ Sentence-BERT \cite{reimers2019sentence} to retrieve the top-$k$ textually similar 3D assets from the Objaverse list based on these descriptions.
From a \textit{visual perspective}, we further validate the selection by using a VLM. 
The VLM takes the task name, detailed task description, and the assets annotations as input.
It then determines whether the scene is valid and outputs "yes" to confirm alignment between the task description and the scene setup.

In this way, we compile a high-quality list of relevant 3D assets that undergo dual validation through both textual and visual steps.

\textbf{Asset Configuration.}
Assets configuration ensures that the retrieved 3D assets have physically plausible dimensions. To automate this process, we employ LLM to determine the size of each asset. 
The size of each object is calculated as a scalar value in meters, representing its largest dimension.
In addition, we establish \textit{spatial relationship rules}, where objects with relative size relationships must satisfy specific task requirements. 
For example, in a scenario where a "bowl of soup" needs to be placed inside a microwave, if the bowl's size is 0.15 meters, the microwave must have a dimension larger than 0.15 meters to accommodate it.
We also define \textit{initial state rules}, ensuring that objects are in the appropriate state for the task. 
For example, for the task "turn off faucet," the faucet must be initialized in the "on" state to accurately simulate the conditions necessary for the task's execution.

\textbf{Scene Configuration.}
After configuring the assets, the final step is to position them accurately while maintaining appropriate spatial relations. 
To achieve precise placement, we query an LLM to establish a 3D world coordinate system $(x, y, z)$. 
Target assets required for specific tasks are strategically placed within the constrained space of $(0, 0, 0)$ to $(1, 1, 1)$, while auxiliary assets not involved in the current task are positioned outside this specified range.

We will demonstrate through human evaluation in \S\ref{solution} that the environment constructed in this manner is of high quality.

\subsection{Proactive Anomaly Detection}
Although existing research extensively explores ways to enhance a robot's ability to follow instructions, a key limitation is their inability to actively detect anomalies or dangerous situations in daily life \cite{fan2024embodied}, an ability that is crucial for ensuring safety in dynamic environments \cite{lundstrom2015holistic,wang2023robogen}.
In our simulated 3D environment, we aim to enable household robots to proactively detect hazards or anomalies and acquire the necessary skills to solve tasks related to these anomalies.
Concretely, \model{} employs an LLM to analyze and identify potential problems.
The input for the LLM includes outputs from the previous steps, including the target object, retrieved assets, their configurations, and the overall scene setup. 
Note that LLM has no access to the task name and description, but only infers the potential task based on the environment observation.
Based on this information, the LLM is prompted to generate possible problems that need to be solved and their solutions.
For example, a problem could be that a folding knife on the table may cause cuts or injuries, and the solution would be to store the folding knife in a storage box.

\label{section:anomaly detection}

\subsection{Anomaly Task Learning}
\label{section::learning}


After confirming the task to be completed, \model{} queries the LLM to decompose the detected solution into shorter-horizon sub-tasks, as illustrated in the bottom part of Figure~\ref{fig:Framework}.

Then, different learning algorithms are selected, tailored to different subtasks: reinforcement learning \cite{schulman2017proximal, haarnoja2023learning}
and action primitives with motion planning \cite{karaman2011sampling}.
Each algorithm has its strengths:
Reinforcement learning is ideal for dynamic, contact-rich environments, like legged locomotion or adjusting appliance controls, and
Action primitives with motion planning handle navigation through cluttered environments, ensuring safe and efficient paths.
We introduce the strengths of each algorithm and provide three examples of action-algorithm pairs to the LLM, enabling it to select the most appropriate learning algorithm for each subtask by in-context learning.


Meanwhile, for subtasks trained using reinforcement learning, the LLM is responsible for generating the corresponding reward functions. 
For tasks involving rigid manipulation and locomotion, these reward functions are derived from low-level states accessible to the LLM. 
In contrast, for tasks involving soft body manipulation, the reward functions are based on the earth-mover distance between the particles of the current and target shapes, ensuring precise shape matching.
To simplify object grasping and approaching action in action primitives subtasks, we use a robot equipped with a suction cup. 
This setup streamlines the process of grasping. 
The simplified pseudo-algorithm for the grasping and approaching primitives is in Appendix \ref{appendix:grasping algorithm}.

\begin{table*}[h]
    \centering
    \scriptsize
    \begin{tabular}{cccccccc}
      \toprule
      & \model{} &RoboGen  & Behavior-100 & RLbench  & MetaWorld &  Maniskill2 & GenSim \\ \midrule
   Number of Tasks  &  111&106   & 100  & 106 & 50 & 20 & 70 \\
    Task Description - Self-BLEU $\downarrow$ & $\mathbf{0.227}$ & 0.287 & 0.299 & 0.317 & 0.322 & 0.674 & 	0.378 \\ 
    
    Task Description -  SentenceBert $\downarrow$ &$\mathbf{0.245}$  & 0.394 & 0.210 & 0.200 & 0.263 & 0.194 & 0.288 \\
    
    Scene Image - Embedding Similarity (ViT) $\downarrow$ & $\mathbf{0.315}$&0.353 & 0.389 & 0.375 & 0.517 & 0.332 & 0.717 \\
    
    Scene Image - Embedding Similarity (CLIP) $\downarrow$ & $\mathbf{0.805}$ & 0.824 & 0.833 & 0.864 & 0.867 & 0.828 & 0.932\\
    \bottomrule
    \end{tabular}
    \caption{\textbf{Task diversity comparison} with leading human-designed robotics datasets, including Behavior-100, RLBench, MetaWorld, Maniskill2, and GenSim.}
    
    \label{tab:diversity}
\end{table*}

\section{Experiment Setup}

\subsection{Implementation Details}

Our proposed system is generic and agnostic to specific simulation platforms.
However, considering the broad audience for simulation platforms, \textcolor{black}{as following \cite{wang2023robogen,kant2022housekeeptidyingvirtualhouseholds,gu2023maniskill2,shridhar2020alfred,savva2019habitat}}. we choose Genesis \cite{katara2023gen2sim}, the most widely employed simulation platform for deployment. 
The model itself is general-purpose and independent of any specific simulation platform. 
We employ the state-of-the-art \texttt{GPT-4-0314} LLM and BLIP-2~\cite{li2023blip} VLM by default.
For anomaly task learning, we utilize the Soft Actor-Critic (SAC) \cite{haarnoja2018soft} as the reinforcement learning algorithm, employing learning rate $3e-4$ for the actor, the critic, and the entropy regularizer. The horizon of manipulation sub-tasks is 100, with a frameskip of 2. For each sub-task, we train with 1M environment steps.  
 We also employ Batch Informed Trees \cite{gammell2015batch} as the motion planning algorithm.
More details are in the Appendix \ref{appendix:Skill learning}.

\subsection{Baselines}

We compare our constructed environment with the latest benchmark environments, including RLBench \cite{james2020rlbench}, which encompasses 100 distinct, meticulously crafted tasks, ranging in complexity from basic tasks like target-reaching and door-opening to more advanced, multi-step tasks.
ManiSkill2 \cite{gu2023maniskill2} includes 20 manipulation task families which cover stationary/mobile-base, single/dual-arm, and rigid/soft-body manipulation tasks with 2D/3D-input data. 
Meta-World \cite{yu2020meta}, serves as a benchmark specifically designed for evaluating the performance of meta-reinforcement learning and multitask learning algorithms. 
Behavior-100 \cite{li2023behavior} featuring 100 activities within simulated environments. 
These activities encompass a variety of routine household tasks, including cleaning, maintenance, and food preparation. 
GenSim \cite{wang2023gen} offers 100 robotic arm grasping scenarios, all set on a tabletop environment.
RoboGen \cite{wang2023robogen} has generated 106 more diverse tasks, further expanding them to accommodate a wider range of robotic arm types. Note that RoboGen doesn't release its constructed dataset; therefore, we reimplement RoboGen based on the provided code.

\subsection{Evaluation Metrics}
We first evaluate the \textit{diversity} of the generated anomaly task settings, including the semantic aspects of the tasks and the visual aspects of the scenes.
For semantic view, we concatenate the task name and description from each anomalous task and calculate their similarity by Self-BLEU \cite{papineni2002bleu} and SentenceBERT \cite{reimers2019sentence} for quantitative analysis. 
For scene visual information, we assess scene diversity by measuring the embedding similarity of unrendered images of the scene from the initial camera state. 
Specifically, we utilize ImageNet pre-trained Vision Transformer (ViT) \cite{dosovitskiy2020image} and CLIP models \cite{radford2021learning}.

Next, we assess whether the robot can proactively and accurately \textit{detect} anomaly scenarios.
We employ three undergraduate students specializing in computer-related fields to determine whether the task detected by the anomaly detection module aligns with the ground truth task.

Finally, the annotators review task action videos to determine whether the task is successfully completed and evaluate the \textit{overall success rate} of task completion.
Note that even if the detected tasks do not fully align with the designed tasks in the task proposal, evaluations are still based on the detected tasks, as these tasks possess a certain degree of rationality, as explained in \S\ref{section:anomaly}.
\begin{figure*}[tb]
    \centering
    \includegraphics[width=\linewidth,height=0.62\linewidth]{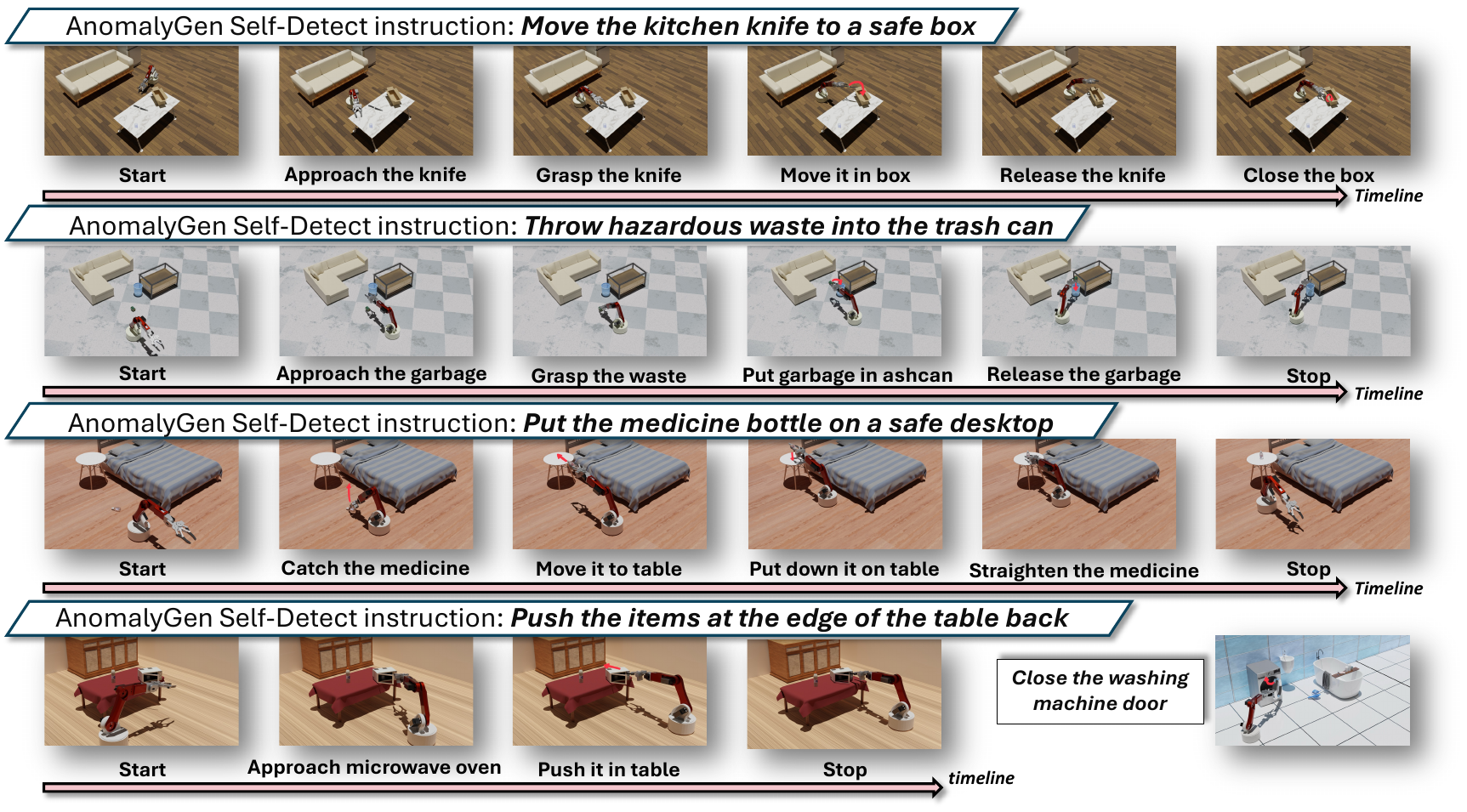}
    \caption{Snapshots of the learned skills across 4 exemplary long-horizon sequential tasks and 1 single-step task.}
    \label{pic::snapshots}
\end{figure*}

\section{Results and Analysis}

\subsection{Anomaly Task Statistics}

\begin{figure}[tb]
    \centering
    \includegraphics[width=0.7\linewidth]{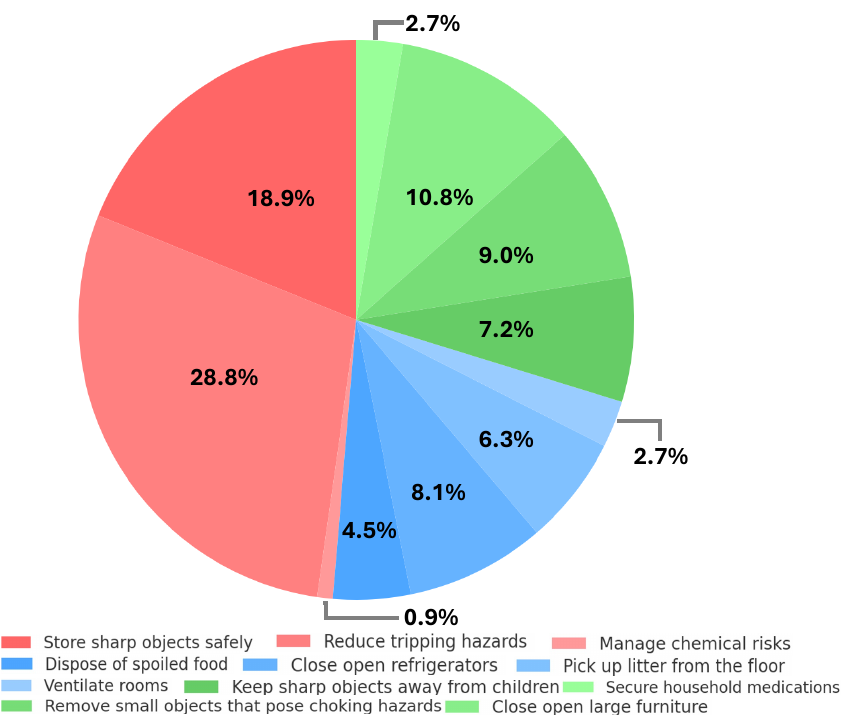}
    \caption{Distribution of types of anomalous scenarios. The red color represents "Household Hazards," the blue color denotes "Hygiene Management," and the green color denotes "Child Safety Measures."}
    \label{fig:stats}
\end{figure}

Through our group brainstorming component, we can theoretically generate an unlimited number of diverse tasks. 
To facilitate the evaluation process, we limited the number of evaluated scenes to 111.
These scenes are categorized into three general categories: household hazards, hygiene management, and child safety. 
Each category includes diverse tasks, as visualized in Figure \ref{fig:stats}, where "store sharp objects safely" represents the largest proportion of tasks at 28.8\%, followed by other critical tasks such as "reduce tripping hazards" and "manage chemical risks," all of which represent common dangers encountered in real life.
Note that no human intervention was involved in the entire design process, so the distribution of tasks is the result of fully autonomous generation by our model.

\subsection{Anomaly Task Diversity}
In Table~\ref{tab:diversity}, we show the diversity of our model and baselines on the constructed anomalies task from both text and visual perspectives. 
It can be seen that our model achieves the lowest Self-BLEU and SentenceBERT similarity scores for text, as well as the lowest ViT and CLIP scores for visuals, indicating that our framework surpasses previous manually constructed benchmarks and datasets.

\textbf{Group Brainstorming Ablation.}
To demonstrate that the diversity described above is due to our group brainstorming setting rather than the LLM's inherent abilities, we conduct an ablation study. 
The first setting removes both the brainstorming and role-play components introduced in \S\ref{brain}, directly querying LLM to generate task proposals.
The second setting incorporates role-playing, where we provide the role settings in the prompt and query LLM to propose tasks from the perspective of the assigned character.
To comprehensively evaluate the generation ability, we generate 300 anomalous task proposals.
Concretely, we use identical query parameters for both experiments and randomly selected 10 assets as targets.
Each method is run 10 times, generating 10 task proposals per iteration. 
We employ Self-BLEU, SentenceBERT, and Word Mover's Distance (WMD)~\cite{huang2016supervised} for evaluation. 

\begin{table}[tb]
    \centering
    \resizebox{0.5\textwidth}{!}{
    \begin{tabular}{lccc}
    \toprule
         Method & Self-BLEU ($\downarrow$) & SentenceBERT ($\downarrow$) & WMD ($\downarrow$) \\
         \midrule
        w/o brainstorming \& role-play & 0.217 & 0.553 & 0.618 \\
        w/o brainstorming & 0.225 & 0.551 & 0.605 \\
        \model{} & \textbf{0.043} & \textbf{0.393} & \textbf{0.511} \\
    \bottomrule
    \end{tabular}}
    \caption{Ablation experiment results on group brainstorming and role-play. Lower values indicate better performance across all metrics.}
    \label{table:ablation}
\end{table}

\begin{figure*}[tb]
    \centering
    \includegraphics[width=1\linewidth]{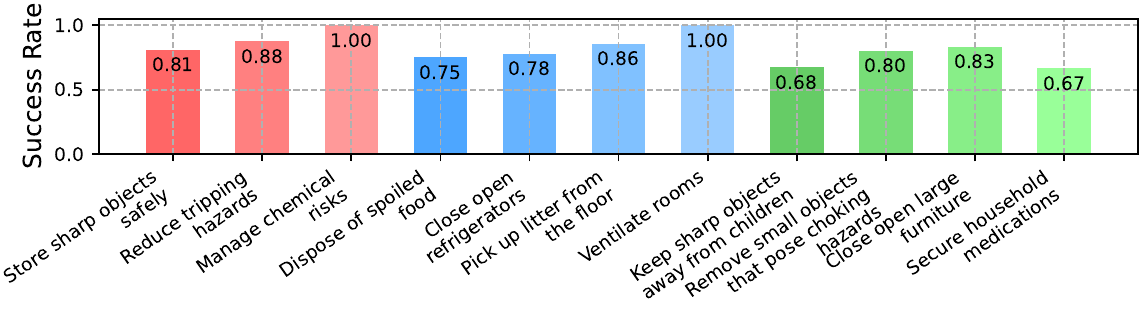}
    \caption{Anomaly resolution completion rate across different categories. }
    \label{fig:resolve}
\end{figure*}

The experiment results are presented in Table~\ref{table:ablation}. 
It can be seen that with role-play, the SentenceBERT and WMD scores decrease by 0.002 and 0.007, respectively, compared to the naive LLM approach. 
Then, brainstorming brings the greatest contribution, leading to the largest improvement across all three metrics, including a 0.182 decrease in Self-BLEU, 0.158 decrease in SentenceBERT, and 0.107 decrease in WMD. 
This demonstrates that the combination of brainstorming and role-play significantly enhances the model's performance by promoting better alignment and understanding, as reflected in the lower values across all metrics.

\subsection{Anomaly Detection Performance}
\label{section:anomaly}
For the anomaly detection task, we allow \model{} to come up with up to three solutions, and manually evaluate the hit accuracy for $k$ solutions. 
As shown in Table \ref{tab:detection_rate}, our model generally achieves high performance, with 76\% accuracy on the first attempt and 82\% accuracy when given three attempts. 
This demonstrates the effectiveness of the prompt we design for \model{} to correctly identify the task, as well as the validity of our constructed simulation environment, where the anomaly scenes can be accurately detected. 

In addition, we conduct an error analysis on the failed tasks.
We find that the main reason for failure was that our environment closely mirrors real-life scenarios, including a variety of everyday clutter, which mislead the LLM into making incorrect detections.
Meanwhile, we note that the tasks proposed by the detection module still possess a degree of rationality. 
For instance, while the ground truth involves picking up a pill from the floor and placing it on a table, our detection module instead proposes to discard it. 
Although this action does not align with the specified ground truth, the alternative task remains meaningful, as discarding a potentially misplaced pill could also be seen as a reasonable safety precaution.

\begin{table}[tb]
\centering    
\scalebox{0.8}{
\begin{tabular}{cc}
\toprule
$k$ Solutions  & Success Rate  \\
\midrule
1  &  0.759 \\ 
2 &  0.804 \\
3&  0.821  \\
\toprule
\end{tabular}
}
\caption{Anomaly detection success rate with $k$ solutions.}
\label{tab:detection_rate}
\end{table}


\subsection{Anomaly Solution Performance}
\label{solution}
Lastly, we can evaluate the performance of overall task performance.
We first give some examples of the task execution process in Figure \ref{pic::snapshots}, including multi-step tasks, such as placing a medicine bottle on a safe desktop, and single-step tasks, such as closing the washing machine door. 
It can be observed that the model successfully follows the instructions and completes tasks of varying complexity and time steps.

Next, we conduct a quantitative analysis of the anomaly resolution accuracy, using human evaluations to assess the completion rate across different categories, as in Figure \ref{fig:resolve}. 
Among the 111 generated anomalous scenarios, the average success rate for resolving these tasks was 83\%, highlighting \model{}'s strong execution performance.

We also conduct an error analysis and summarize two main reasons for task failure. 
First, there is an overlap of scene assets. 
In the verification step described in \S\ref{verification}, we employ VLMs to verify that the assets are correctly positioned. 
However, there are 4 out of 111 instances where the VLMs fail to identify mispositioned items and incorrectly approve them. 
For example, when the mispositioned items are both white, accurate identification becomes more difficult. 
The misposition problem exists in other environments~\cite{wang2023robogen}, and we anticipate that advancements in VLM technology will address this limitation.
Second, some tasks proposed by \model{} include complex, multi-step actions that challenge the capabilities of current algorithms, making it difficult for them to perform effectively. 
We expect that improvements in learning algorithms will enable robots to better learn from their environments and handle more complex tasks in the future.

\section{Conclusion}
In this study, we present \model{}, an innovative framework designed to enhance the proactive detection and resolution of household anomalies by robots. 
Our approach integrates advanced generative models to automatically create diverse and realistic 3D environments, which are essential for training robots to handle real-world tasks autonomously.
We also propose a group brainstorming method, which generates a wide variety of anomalous scenarios, surpassing traditional methods that rely heavily on manual input. 
Furthermore, the \model{} framework introduces a novel approach to anomaly detection, offering potential strategies for enabling robots to act without direct human commands. 
We hope our work will inspire further exploration into autonomous decision-making in real-world applications.


\section{Limitation}
While \model{} has achieved certain milestones, it still encounters several limitations:

    1) In unsupervised settings, the validation of tasks within generated anomaly scenarios remains challenging, with a potential for scenarios that clearly do not meet task requirements. This issue is particularly exacerbated under conditions of large-scale generation. However, with future enhancements in the capabilities of multimodal large language models, we anticipate that this limitation will be addressed.  
    
    2) The richness of the generated scenes is currently somewhat constrained by the scale of the 3D assets dataset. A limited dataset size may curtail the full potential of \model{}.  
    
    3) Regarding the deployment of \model{} into real-world applications, there remains a significant sim-to-real domain gap. This gap constitutes an independent research domain that is beyond the scope of our current work. Given recent rapid advancements in physically accurate simulation \cite{li2020incremental} and techniques such as domain adaptation \cite{tobin2017domain,xu2023roboninja,10507724} along with realistic sensory signal rendering \cite{zhuang2023robot}, we anticipate a continual narrowing of this gap in the near future.

\bibliography{custom}

\appendix

\section{3D Assets Stats}
We have compiled statistics regarding the types and quantities of 3D assets extracted from PartNet-Mobility. The details are presented in Table \ref{app:table:subet}.
In summary, our subset comprises 44 categories of 3D assets and a total of 2,193 individual 3D assets.
\label{appendix: assets stats}
\begin{table*}[]  
\centering  
\begin{tabular}{c|c|c|c}  
\toprule  
Type of item & Number of models & Type of item& Number of models \\ \hline   
Bottle & 57 & Microwave & 16 \\ \hline  
Box & 28 & Mouse & 14 \\ \hline  
Bucket & 36 & Oven & 30 \\ \hline  
Camera & 37 & Pen & 48 \\ \hline  
Cart & 61 & Phone & 18 \\ \hline  
Chair & 81 & Pliers & 25 \\ \hline  
Clock & 31 & Printer & 29 \\ \hline  
CoffeeMachine & 54 & Refrigerator & 44 \\ \hline  
Dishwasher & 48 & Remote & 49 \\ \hline  
Dispenser & 57 & Safe & 30 \\ \hline  
Display & 37 & Scissors & 47 \\ \hline  
Door & 36 & Stapler & 23 \\ \hline  
Eyeglasses & 65 & StorageFurniture & 346 \\ \hline  
Fan & 81 & Suitcase & 24 \\ \hline   
FoldingChair & 26 & Table & 101 \\ \hline  
Globe & 61 & Toaster & 25 \\ \hline  
Kettle & 29 & Toilet & 69 \\ \hline  
Keyboard & 37 & TrashCan & 70 \\ \hline  
KitchenPot & 25 & USB & 51 \\ \hline  
Knife & 44 & WashingMachine & 17 \\ \hline  
Lamp & 45 & Window & 58 \\ \hline  
Laptop & 55 & Lighter & 28 \\ \bottomrule  
\end{tabular}  
\caption{ Detail categories and quantities of subset which select from PartNet-Mobility. }\label{app:table:subet}    
\end{table*}

\section{Brainstorming Setting}

\subsection{Role List}
\label{appenix::Role list}
In this section, we provide a detailed list of roles that are commonly found within a household setting, each accompanied by a specific role description. These roles encompass a variety of responsibilities and skills required to efficiently manage and maintain a home environment. Table \ref{app:tab:role-list} outlines the diverse roles ranging from daily household management to specialized services that enhance the functionality and comfort of home life.

\begin{table*}[h]
\centering
\begin{tabular}{>{\raggedright\arraybackslash}p{4cm}|p{10cm}}
\toprule
\textbf{Role} & \textbf{Role Description} \\
\hline
Homemaker & Responsible for managing household chores and daily life, acting as the heart of the home. Skills include expert cooking, time management, and budget control. The challenge lies in providing the best quality of life on a limited budget. \\
\hline
 Engineer & Specializes in designing and maintaining home-use robots such as cleaning robots or elder care robots. Skills in programming, mechanical design, and AI. The challenge is developing robots that integrate seamlessly into the home environment. \\
\hline
Gardener & In charge of designing and maintaining the home garden. Knowledge in botany, creative design, and ecological maintenance. The challenge is to create an aesthetically pleasing yet sustainable outdoor space. \\
\hline
Nutritionist & Provides dietary advice and plans for family members. Expertise in nutrition, food science, and health promotion. The challenge is to balance various dietary restrictions and preferences. \\
\hline
Personal Trainer & Responsible for physical training and health management of family members. Skills in sports science, human physiology, and motivational psychology. The challenge is to create personalized fitness programs that accommodate varying fitness levels. \\
\hline
Financial Planner & Manages family finances, providing investment and savings advice. Knowledge in economics, market analysis, and risk management. The challenge is ensuring financial security and future growth for the family. \\
\hline
Educational Consultant & Supports children in the family with academic guidance and educational planning. Expertise in pedagogy, psychology, and curriculum design. The challenge is to adapt to different learning styles and educational needs. \\
\hline
Home Security Officer & Responsible for family safety and handling emergencies. Skills in security management, emergency response, and physical defense. The challenge is to maintain security without compromising the family's freedom and comfort. \\
\hline
Interior Designer & Optimizes the layout and design of the home to enhance living experience. Skills in artistic design, spatial planning, and color theory. The challenge is to create a functional and beautiful living space within budget. \\
\hline
Household Advisor & Provides comprehensive home management services from daily cleaning to organizing special events. Skills in project management, customer service, and efficiency optimization. The challenge is ensuring all household activities run efficiently and seamlessly. \\
\bottomrule
\end{tabular}
\caption{Roles and Descriptions for Household-Based Role Play}\label{app:tab:role-list}
\end{table*}


\subsection{Detail Prompt}
\begin{figure*}
     \centering
     \includegraphics[width=1\linewidth]{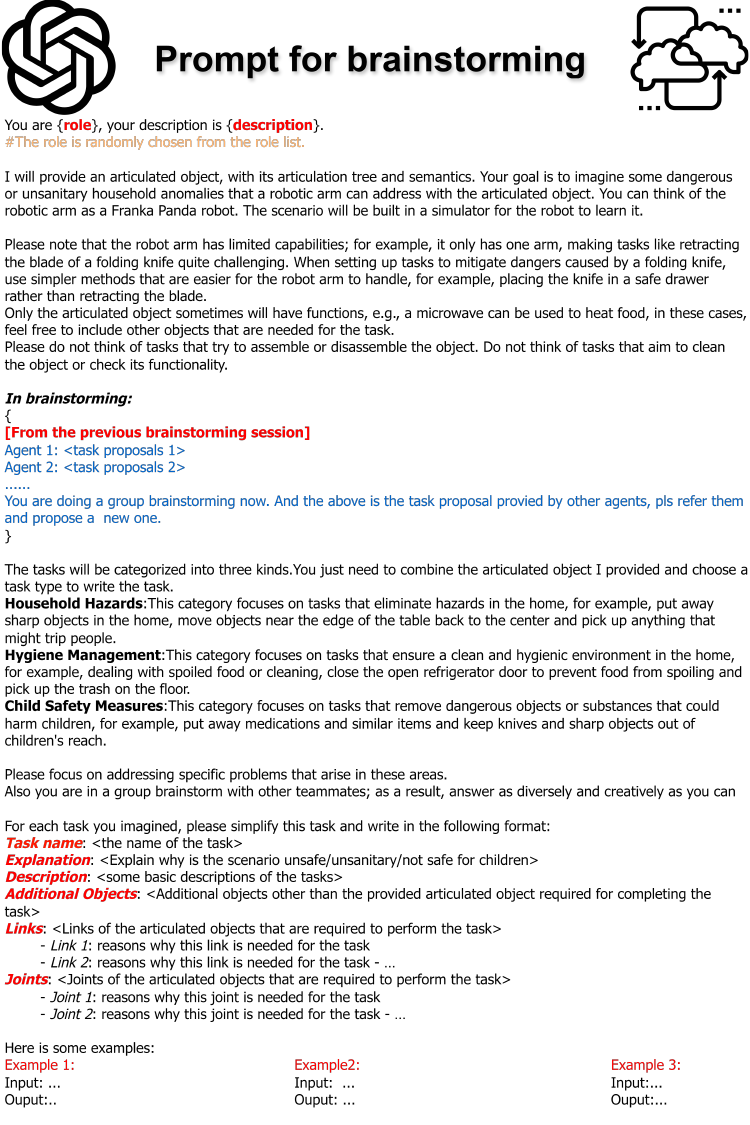}
     \caption{The prompt template of Brainstorming.}
     \label{fig:brainstorming}
 \end{figure*}

We show our prompt template in Figure \ref{fig:brainstorming}.
The prompt outlines a brainstorming session focused on generating home safety tasks for the Franka Panda robotic arm, taking into account the articulated, semantically tagged movable objects in a household setting. These tasks are to be envisioned in scenarios that may pose potential hazards or unsanitary conditions within the home, which the robot is equipped to handle. The tasks are categorized into three primary areas: household hazard, hygiene management, and child safety measures. Each task is to be formatted to include the task name, an explanation, a description, any auxiliary items required, and the articulations and their specific functions. The brainstorming context should be collaborative, with a strong emphasis on the operational limits of the robotic arm, such as avoiding complex assembly, disassembly, or cleaning tasks. This ensures that the tasks are tailored to the robot’s capabilities, focusing on practical and manageable interventions in household environment.

\label{appendix: detail prompt}

\section{Parameter Setting}

\subsection{Algorithm of Grasping and Approaching Primitives}
\label{appendix:grasping algorithm}
\begin{algorithm}\small
\caption{Grasping and Approaching Primitives}\label{alg::grasping}
\begin{algorithmic}[1]
\State \textbf{Initialize:}
\State  \quad TargetObject: the object or link to be manipulated
\State  \quad GripperPose: the pose of the robotic gripper
\State \quad \textbf{Procedure:}
\State \quad Point $p \gets \text{RandomSample(TargetObject)}$
\State \quad Vector $n \gets \text{NormalAtPoint(TargetObject, } p)$
\State \quad GripperPose $\gets \text{AlignWithNormal(GripperPose, } n)$
\State \quad Path path $\gets \text{MotionPlanning(GripperPose)}$
\State \quad \text{ExecutePath(path)}
\State \quad \textbf{while} \text{not ContactMade()} \textbf{do}
\State \quad \quad \text{MoveAlongNormal(GripperPose, } n$)$
\State \quad \textbf{end while}
\State \quad \textbf{if} \text{ContactMade()} \textbf{then}
\State \quad \quad \text{Grasp(TargetObject)}
\State \quad \textbf{end if}
\end{algorithmic}
\end{algorithm}

In designing a robotic manipulator equipped with a suction cup to facilitate object grasping, the operational primitives for grasping and approaching are outlined as follows: Initially, a random point on the surface of the designated target object or link is selected. Subsequently, a gripper pose is calculated such that it aligns with the normal at the sampled point. Motion planning algorithms are then employed to devise a collision-free trajectory to the predetermined gripper pose. Upon attaining this pose, the manipulator advances along the normal vector until contact is established with the object. In this setup, \model{} leverages LLM to determine the specific target object for either grasping or approaching, dependent on the given subtask. We show the simplified pseudo-algorithm in Algorithm \ref{alg::grasping}.
\subsection{Skill Learning Parameter}
\label{appendix:Skill learning}

For anomaly task learning, we employ SAC algorithm for reinforcement learning. In object manipulation tasks, the observation space includes the low-level state of objects and the robot involved in the task. The SAC utilizes MLP with three layers, each having 256 units, for both the policy and Q networks. We set a learning rate of 3e-4 for the actor, critic, and entropy regularizer. Each manipulation task has a horizon of 100 steps and employs a frameskip of 2. The RL policy controls a 6-dimensional action space, where the first three dimensions dictate the translation—either as delta translation or target location—and the remaining three dimensions specify the delta rotation, represented as a delta-axis angle in the gripper’s local frame. We train each sub-task over 1 million environment steps.

For locomotion tasks, we apply the Cross Entropy Method (CEM) for skill learning, which has proven more stable and efficient than traditional RL approaches. We use a ground-truth simulator as the dynamics model in CEM, focusing on optimizing the joint angle values of the robot. The horizon for locomotion tasks is set to 150, with frameskip of 4.

Additionally, we integrate BIT \cite{gammell2015batch}, implemented within the Open Motion Planning Library OMPL \cite{sucan2012open}, for action primitives in motion planning. Specifically, for grasping and approaching primitives, we begin by sampling a surface point on the targeted object or link. We then compute a gripper pose that aligns the gripper’s y-axis with the normal of the sampled point. The pre-contact gripper pose is established 0.03 meters above the surface point along the normal direction. Utilizing motion planning, we identify a collision-free path to the target gripper pose, continuing the gripper’s movement along the normal until contact is achieved.

\section{Data Statistics}
We present the categories and numbers of all generated scenes. Detailed statistics are in Table~\ref{table:stats}.
\begin{table*}[h!]
\centering
\resizebox{\textwidth}{!}{
\begin{tabular}{l|c|c}
\toprule
\textbf{Category} & \textbf{Class Name} & \textbf{Number} \\
\midrule
\multirow{3}{*}{Household Hazards} & Store sharp objects safely & 21 \\ 
                                   & Reduce tripping hazards &  32\\ 
                                   & Manage chemical risks & 1  \\ 
\multirow{4}{*}{Hygiene Management} & Dispose of spoiled food & 5 \\ 
                                    & Close open refrigerators & 9 \\ 
                                    & Pick up litter from the floor &  7\\ 
                                    & Ventilate rooms & 3 \\ 
\multirow{4}{*}{Child Safety Measures} & Keep sharp objects away from children & 8 \\ 
                                       & Remove objects that pose choking hazards & 10 \\ 
                                       & Close open large furniture (cabinets, dishwashers, etc.) & 12 \\ 
                                       & Secure household medications & 3 \\ 
\bottomrule
\end{tabular}
}
\caption{Household Hazards, Hygiene Management, and Child Safety Measures}\label{table:stats}
\end{table*}

\end{document}